\documentclass[runningheads]{llncs}
\usepackage[T1]{fontenc}

\usepackage{graphicx,verbatim}
\usepackage{amsmath}
\usepackage{amssymb}
\usepackage{bbm}
\begin{document}
\title{SecondOpinion: Anatomy-Aware Gated Reasoning for Efficient Medical Image Analysis}
\titlerunning{SecondOpinion}

\author{Siam Tahsin Bhuiyan\inst{1, 2}\orcidID{0009-0000-1298-1991}, Rashedur Rahman\inst{1, 2}\thanks{Corresponding author}\orcidID{0000-0003-0267-2612}, Sefatul Wasi\inst{2}\orcidID{0009-0004-6949-0607}, Riyadul Islam\inst{1}\orcidID{0009-0007-8702-9330}, Syoji Kobashi\inst{3}\orcidID{0000-0003-3659-4114}, Ashraful Islam\inst{1, 2}\orcidID{0000-0003-2367-2013}, Saadia Binte Alam\inst{1, 2}\orcidID{0009-0007-0358-7635}}


\authorrunning{S. T. Bhuiyan et al.}
\institute{Center for Computational \& Data Sciences, Independent University, Bangladesh, Dhaka, Bangladesh \and
Department of Computer Science and Engineering, Independent University, Bangladesh, Dhaka, Bangladesh \and
Graduate School of Engineering, University of Hyogo, Kobe, Japan \\
\email{rashed@iub.edu.bd}}

\maketitle

\begin{center}
\small\itshape
Preprint. Accepted at EMA4MICCAI 2026 (MICCAI Workshop).
\end{center}

\vspace{0.5em}
\begin{abstract}
Deep learning models for medical image analysis typically apply a fixed amount of computation to every input, regardless of case difficulty. Anatomy-guided dual-stream architectures have been shown to improve diagnostic performance, but they evaluate both streams unconditionally, even on cases a single stream could already resolve confidently. We propose SecondOpinion, a framework in which a fast primary stream processes every case, while a second, anatomy-guided stream is invoked only when GateKeeper, a gating mechanism trained explicitly as a binary correctness classifier, judges that the primary stream's prediction needs additional scrutiny, much as a clinician might seek a second opinion on a difficult case. When activated, the two streams are combined through a lightweight cross-attention fusion module. We evaluate SecondOpinion on a unified five-class chest X-ray dataset and a pelvic fracture dataset, the latter including a held-out, harder subset of fractures that are invisible on X-ray but confirmed via CT. SecondOpinion matches or exceeds prior state-of-the-art performance on both tasks, while activating its anatomy-guided stream on only 9.23\% of chest X-ray cases, rising to 24.12\% on visible fractures and 45.71\% on invisible fractures, an activation rate that tracks task difficulty directly. These results suggest that supervising a gating signal toward correctness, rather than relying on unsupervised confidence, allows a model to allocate anatomical reasoning where it is actually needed.
\end{abstract}

\keywords{Medical Image Analysis  \and Conditional Computation \and Anatomy-guided Learning \and Dual-stream Networks \and Chest X-ray Classification\and Fracture Detection.}
\section{Introduction}
Computer-aided diagnosis from medical images has advanced rapidly, with deep learning models achieving expert-comparable performance across diverse imaging modalities and pathologies \cite{rajpurkar2017chexnet}. Most diagnostic models, however, apply a fixed amount of computation to every input regardless of case difficulty: an obvious finding is processed at the same cost as an ambiguous one. In clinical practice, diagnostic effort is naturally adaptive, with straightforward cases resolved quickly while ambiguous ones prompt a clinician to seek a second opinion. This motivates our central question: can a diagnostic model learn to allocate additional anatomical reasoning only when it is actually needed?

Anatomy-guided deep learning has been shown to improve recognition performance, particularly for spatially sparse or low-contrast structures \cite{jin2022anatomy}, by infusing anatomic knowledge into the network design itself rather than leaving it to be discovered unsupervised \cite{bhuiyan2025invisible}. PelFANet \cite{bhuiyan2025invisible} follows this direction for pelvic fracture detection, using anatomy-aware feature guidance to improve sensitivity to subtle fractures. Dual-stream architectures are a common mechanism for incorporating such guidance, processing anatomical and pathological information through separate representations before fusion \cite{zafar2025dual,bruno2025dual}. A consistent limitation across this line of work is that both streams are evaluated on every input unconditionally, doubling inference cost even for the large fraction of cases a single stream could already resolve confidently.

A separate line of work addresses efficiency directly through conditional computation, adapting computational effort to each input rather than reducing cost uniformly for all inputs \cite{passalis2020efficient}. Early-exit networks are the most established instance of this idea, using a gating mechanism to decide whether further computation can be skipped once a confident prediction is reached \cite{wolczyk2021zero,mokssit2025confidence}. Recent work shows that aligning training dynamics with the inference-time gating policy improves early-exit reliability \cite{mokssit2025confidence}, motivating our use of explicit, ground-truth-supervised confidence signals rather than relying on implicit task-loss gradients alone.

While confidence-gated dual-path designs exist in retrieval-augmented classification \cite{tang2026t}, correctness-supervised gating over anatomy-guided streams has not been explored. We bridge this gap with SecondOpinion: a fast primary stream handles every case, and an anatomy-guided second stream is invoked only when GateKeeper, a learned gating mechanism, flags low confidence in the primary stream's prediction, much as a clinician might seek a second opinion on a difficult case.

Our contributions are as follows:
\begin{enumerate}
    \item We propose SecondOpinion, an adaptive dual-stream framework in which an anatomy-guided "second opinion" stream is invoked conditionally, rather than on every input as in prior dual-stream designs.
    \item We introduce GateKeeper, a lightweight gating mechanism trained explicitly as a binary correctness classifier, predicting the probability that the primary stream's prediction is correct, rather than relying on an unsupervised or heuristically thresholded confidence score.
    \item We design a lightweight cross-attention fusion module, invoked only when GateKeeper activates the second stream, that allows the primary stream's representation to selectively attend to anatomical context.
    \item We validate SecondOpinion on two datasets of increasing diagnostic difficulty: multi-class chest disease detection (COVID-19, viral pneumonia, lung opacity, tuberculosis, normal) and pelvic fracture detection, generalizing from visible to invisible fractures. GateKeeper's activation rate tracks this difficulty gradient (9.23\% to 24.12\% to 45.71\%), while matching or exceeding prior state-of-the-art at a fraction of the always-on computational cost.
\end{enumerate}

\section{Methodology}
\begin{figure}
\includegraphics[width=\textwidth]{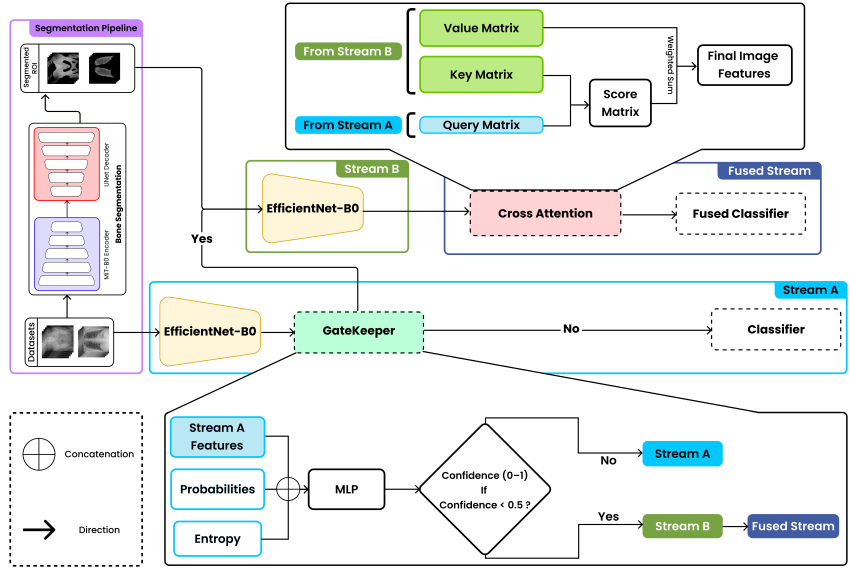}
\caption{SecondOpinion Full Pipeline.} \label{fig:architecture}
\end{figure}

SecondOpinion processes every input through a fast primary stream, and invokes a second, anatomy-guided stream only when a learned gating mechanism, GateKeeper, flags the primary stream's prediction as uncertain. Figure~\ref{fig:architecture} illustrates the full pipeline: a primary stream (Stream A) on the raw image, an anatomy-guided stream (Stream B) on a segmented representation, GateKeeper, and a cross-attention fusion module invoked only when Stream B is activated.

\subsection{Stream A: Primary Pathway}

Stream A processes the raw input $x_a \in \mathbb{R}^{3 \times 224 \times 224}$ through an ImageNet-pretrained EfficientNet-B0 backbone \cite{tan2019efficientnet}, producing a pooled feature vector $h_a \in \mathbb{R}^{1280}$ and logits $z_a \in \mathbb{R}^{C}$ via a linear head, where $C$ is the number of classes. Stream A is always evaluated and constitutes the primary diagnostic pathway.

\subsection{Stream B: Anatomy-Guided Pathway}

Stream B processes an anatomy-guided representation $x_b$, obtained from a U-Net (MiT-B0 backbone) segmentation model trained on ground-truth masks for all classes except tuberculosis, where masks were instead produced via inference; in all cases, Stream B's input is the predicted segmentation, not ground truth. This representation is passed through a second, independently weighted EfficientNet-B0 backbone, producing a pooled feature vector $h_b \in \mathbb{R}^{1280}$ and spatial map $S_b \in \mathbb{R}^{1280 \times 7 \times 7}$. Stream B shares no intermediate features with Stream A, reflecting the clinical analogy that a second opinion should be formed independently of the first.

\subsection{GateKeeper}

GateKeeper estimates $P(\text{Stream A is correct})$ from Stream A's pooled features $h_a$, softmax probabilities $p_a = \text{softmax}(z_a)$, and the Shannon entropy \cite{shannon1948mathematical} of $p_a$,

\begin{equation}
H(p_a) = -\sum_{i=1}^{C} p_{a,i} \log(p_{a,i} + \epsilon)
\label{eq:entropy}
\end{equation}

concatenated and passed through a small MLP $\phi(\cdot)$ to produce 

\begin{equation}
\alpha = \sigma(\phi([h_a; p_a; H(p_a)])) \in (0,1)
\label{eq:gatekeeper}
\end{equation}

GateKeeper activates Stream B whenever $\alpha < 0.5$, the natural decision boundary of the binary cross-entropy objective used to train it, rather than a tuned threshold. This objective targets correctness prediction, not efficiency directly; efficiency follows as a consequence, since a well-calibrated gate activates Stream B only where Stream A is unreliable.

\subsection{Cross-Attention Fusion}

When activated, Stream B's spatial features are fused with Stream A's via single-head cross-attention \cite{vaswani2017attention}, with Stream A as query and Stream B as key/value, reflecting Stream A's role as the primary reasoner attending selectively to anatomical context. Given flattened, projected features $Q = S_a W_Q$, $K = S_b W_K$, $V = S_b W_V \in \mathbb{R}^{N \times d}$ ($N{=}49$ tokens), the fused output is

\begin{equation}
\tilde{F} = \text{softmax}\!\left(\frac{QK^{\top}}{\sqrt{d}}\right) V
\label{eq:crossattn}
\end{equation}

$\tilde{F}$ is mean-pooled and layer-normalized to $\tilde{f} \in \mathbb{R}^d$, concatenated with $h_a, h_b$, and passed through a classification head to produce $z_{\text{fused}} \in \mathbb{R}^{C}$.

\subsection{Training Procedure}
\label{sec:training}

Training proceeds in two phases. In \textbf{Phase 1}, both streams run on every sample with GateKeeper frozen (no reliable correctness signal exists yet), optimizing

\begin{equation}
\mathcal{L}_{\text{phase1}} = \mathcal{L}_{\text{CE}}(z_a, y) + \mathcal{L}_{\text{CE}}(z_{\text{fused}}, y)
\label{eq:phase1}
\end{equation}

In \textbf{Phase 2}, GateKeeper is unfrozen and trained jointly. A soft-gated prediction $z_{\text{final}} = \alpha z_a + (1-\alpha) z_{\text{fused}}$ preserves gradient flow, and GateKeeper is supervised with binary cross-entropy against $\hat{g} = \mathbf{1}[\arg\max(z_a) = y]$, giving the unweighted four-term objective

\begin{equation}
\mathcal{L}_{\text{phase2}} = \mathcal{L}_{\text{CE}}(z_{\text{final}}, y) + \mathcal{L}_{\text{CE}}(z_a, y) + \mathcal{L}_{\text{CE}}(z_{\text{fused}}, y) + \text{BCE}(\alpha, \hat{g})
\label{eq:phase2}
\end{equation}

The two auxiliary terms keep both streams learning even when $\alpha$ saturates toward 0 or 1; no efficiency-penalty term is needed, since low activation on correctly-handled cases follows directly from the gate's supervision objective. All terms are unweighted to avoid additional tunable hyperparameters. At inference, hard gating replaces Equation~\ref{eq:phase2}'s soft combination: Stream B and fusion run only when $\alpha < 0.5$.

\section{Experiments}

\subsection{Datasets}

\textbf{Chest X-ray.} We combine two public sources, the COVID-19 Radiography Database \cite{chowdhury2020can,rahman2021exploring} and the Tuberculosis Chest X-ray Database \cite{rahman2020reliable}, into a unified five-class dataset of 21{,}865 images (10{,}192 normal, 6{,}012 lung opacity, 1{,}345 viral pneumonia, 3{,}616 COVID-19, 700 tuberculosis), resized to $224 \times 224$. Tuberculosis lacked segmentation ground truth; its anatomy-guided input was inferred from the model trained on the other four classes (Section~2.2). We use stratified 5-fold cross-validation with no augmentation.

\textbf{Pelvic Fracture.} We use the private AMERI PXR dataset (228 images: 168 fracture, 60 normal) from the Steel Memorial Hirohata Hospital in Japan, evaluated under stratified 5-fold cross-validation with expert-annotated bone segmentation. A held-out Invisible Fracture (INVIS) subset of 35 cases, confirmed via 3D-CT but not visible on X-ray, evaluates generalization only. Class imbalance was addressed via augmentation (2$\times$ fracture, 6$\times$ normal) using random rotation, shear, flip, and translation.

\subsection{Implementation Details}

All models were trained on a single NVIDIA A100 GPU (Google Colab) with AdamW and CosineAnnealing scheduling, batch size 16, learning rate $1\times10^{-4}$ for both phases. Chest X-ray training used 25/35 epochs (Phase 1/2); pelvic fracture used 15/25 epochs, reflecting dataset scale. All other settings were held constant across tasks.

\subsection{Evaluation Metrics}

We report Accuracy, Precision, Recall, Specificity, F1, and AUROC with 95\% CIs ($z$-approximation) over 5-fold CV. For pelvic fracture, F1 is the harmonic mean of Recall and Specificity, matching the convention used by PelFANet \cite{bhuiyan2025invisible} and the ImageNet/DRR20 baselines \cite{rahman2024enhancing} for direct comparability; chest X-ray F1 uses the standard Precision-Recall form. We also report Params and FLOPs as static per-path values and activation-weighted averages.

\section{Results \& Discussion}

\subsection{Main Results}

Stream A only and Stream B only share identical architectures, differing only in input (raw image versus segmentation-guided representation), and Stream B only consistently outperforms Stream A only across all tasks, confirming that anatomy-guided inputs provide independent diagnostic value.

\textbf{Chest X-ray.} SecondOpinion reaches 98.41\% accuracy and 0.990 AUROC (Table~\ref{tab:cxr}), exceeding CheXNet-CBAM \cite{bhuiyan2026chexnet} while activating Stream B on only 9.23\% of samples, trailing the always-on configuration by just 0.6 accuracy points. We view this gap as the explicit cost of efficiency: SecondOpinion does not aim to outperform an unconditional dual-stream model, but to approximate it closely while avoiding unnecessary computation on cases that do not require it.

\begin{table}[t]
\footnotesize
\setlength{\tabcolsep}{3.5pt}
\renewcommand{\arraystretch}{0.92}
\centering
\caption{Chest X-ray results. Mean over 5-fold CV; 95\% CI shown in the row below each of our models' results.}
\label{tab:cxr}
\begin{tabular*}{\columnwidth}{@{\extracolsep{\fill}}lcccccc@{}}
\hline
\textbf{Model} & \textbf{Acc.} & \textbf{Prec.} & \textbf{Rec.} & \textbf{Spec.} & \textbf{F1} & \textbf{AUROC} \\
\hline
EfficientNetB3 \cite{bhuiyan2025lung} & 93.35\% & 95.99\% & 93.81\% & 95.11\% & 0.949 & 0.987 \\
CheXNet \cite{bhuiyan2025lung} & 94.12\% & 94.76\% & 93.05\% & 95.81\% & 0.939 & 0.988 \\
CheXNet-CBAM \cite{bhuiyan2026chexnet} & 96.21\% & 96.14\% & 94.67\% & 97.55\% & 0.954 & 0.989 \\
\hline
Stream A only & 93.25\% & 95.02\% & 93.56\% & 95.01\% & 0.943 & 0.981 \\
 & \footnotesize$\pm$0.32\% & \footnotesize$\pm$0.40\% & \footnotesize$\pm$0.53\% & \footnotesize$\pm$0.52\% & \footnotesize$\pm$0.0044 & \footnotesize$\pm$0.0061 \\
Stream B only & 94.06\% & 95.53\% & 94.28\% & 95.56\% & 0.949 & 0.983 \\
 & \footnotesize$\pm$0.19\% & \footnotesize$\pm$0.22\% & \footnotesize$\pm$0.61\% & \footnotesize$\pm$0.58\% & \footnotesize$\pm$0.0040 & \footnotesize$\pm$0.0042 \\
Always-On & 99.05\% & 98.56\% & 98.32\% & 99.42\% & 0.984 & 0.993 \\
 & \footnotesize$\pm$0.06\% & \footnotesize$\pm$0.16\% & \footnotesize$\pm$0.24\% & \footnotesize$\pm$0.02\% & \footnotesize$\pm$0.0018 & \footnotesize$\pm$0.0054 \\
\textbf{SecondOpinion} & \textbf{98.41\%} & \textbf{97.41\%} & \textbf{97.13\%} & \textbf{98.91\%} & \textbf{0.973} & \textbf{0.990} \\
 & \footnotesize$\pm$0.07\% & \footnotesize$\pm$0.26\% & \footnotesize$\pm$0.08\% & \footnotesize$\pm$0.02\% & \footnotesize$\pm$0.0013 & \footnotesize$\pm$0.0046 \\
\hline
\end{tabular*}
\end{table}

\begin{table}[!htb]
\footnotesize
\setlength{\tabcolsep}{3.5pt}
\renewcommand{\arraystretch}{0.92}
\centering
\caption{Visible fracture (VIS) results. Mean over 5-fold CV; 95\% CI shown in the row below each of our models' results.}
\label{tab:vis}
\resizebox{\columnwidth}{!}{%
\begin{tabular*}{\columnwidth}{@{\extracolsep{\fill}}lcccccc@{}}
\hline
\textbf{Method} & \textbf{Acc.} & \textbf{Prec.} & \textbf{Rec.} & \textbf{Spec.} & \textbf{F1} & \textbf{AUROC} \\
\hline
ImageNet+DRR20 \cite{rahman2024enhancing} & -- & -- & -- & -- & 0.8390 & 0.9280 \\
DRR20 \cite{rahman2024enhancing} & -- & -- & -- & -- & 0.8520 & 0.9290 \\
PelFANet \cite{bhuiyan2025invisible} & 88.68\% & 92.49\% & 92.21\% & 78.33\% & 0.8471 & 0.9334 \\
\hline
Stream A only & 86.40\% & 85.90\% & 84.00\% & 73.30\% & 0.763 & 0.867 \\
 & \scriptsize{$\pm$3.8\%} & \scriptsize{$\pm$4.3\%} & \scriptsize{$\pm$9.0\%} & \scriptsize{$\pm$20.3\%} & \scriptsize{$\pm$0.121} & \scriptsize{$\pm$0.020} \\
Stream B only & 86.60\% & 86.20\% & 85.80\% & 76.60\% & 0.806 & 0.871 \\
 & \scriptsize{$\pm$3.6\%} & \scriptsize{$\pm$4.2\%} & \scriptsize{$\pm$7.8\%} & \scriptsize{$\pm$9.6\%} & \scriptsize{$\pm$0.071} & \scriptsize{$\pm$0.008} \\
Always-On & 92.60\% & 93.90\% & 89.80\% & 90.70\% & 0.901 & 0.951 \\
 & \scriptsize{$\pm$2.7\%} & \scriptsize{$\pm$0.9\%} & \scriptsize{$\pm$1.9\%} & \scriptsize{$\pm$6.4\%} & \scriptsize{$\pm$0.027} & \scriptsize{$\pm$0.006} \\
\textbf{SecondOpinion} & \textbf{91.40\%} & \textbf{93.00\%} & \textbf{87.90\%} & \textbf{88.30\%} & \textbf{0.879} & \textbf{0.949} \\
 & \scriptsize{$\pm$3.5\%} & \scriptsize{$\pm$1.5\%} & \scriptsize{$\pm$2.9\%} & \scriptsize{$\pm$6.5\%} & \scriptsize{$\pm$0.022} & \scriptsize{$\pm$0.011} \\
\hline
\end{tabular*}%
}
\end{table}

\textbf{Pelvic Fracture.} SecondOpinion outperforms PelFANet \cite{bhuiyan2025invisible} on both subsets in AUROC and F1 (Tables~\ref{tab:vis}, \ref{tab:invis}), trailing only on recall. This is a trade-off: the recall gap is offset by a larger gain in specificity, yielding higher F1 overall. Stream B activates on 24.12\% of visible and 45.71\% of invisible cases, the latter reflecting generalization to unseen, harder cases, with the always-on gap widening to 0.8 accuracy points at this higher activation rate.

\begin{table}[!t]
\footnotesize
\setlength{\tabcolsep}{3.5pt}
\renewcommand{\arraystretch}{0.92}
\centering
\caption{Invisible fracture (INVIS) results, evaluated as a held-out generalization set. Mean over 5-fold CV; 95\% CI shown in the row below each of our models' results.}
\label{tab:invis}
\resizebox{\columnwidth}{!}{%
\begin{tabular*}{\columnwidth}{@{\extracolsep{\fill}}lcccccc@{}}
\hline
\textbf{Method} & \textbf{Acc.} & \textbf{Prec.} & \textbf{Rec.} & \textbf{Spec.} & \textbf{F1} & \textbf{AUROC} \\
\hline
ImageNet+DRR20 \cite{rahman2024enhancing} & -- & -- & -- & -- & 0.7210 & 0.7140 \\
DRR20 \cite{rahman2024enhancing} & -- & -- & -- & -- & 0.7860 & 0.8002 \\
PelFANet \cite{bhuiyan2025invisible} & 82.29\% & 88.36\% & 84.35\% & 78.33\% & 0.8123 & 0.8688 \\
\hline
Stream A only & 76.60\% & 77.90\% & 82.10\% & 71.60\% & 0.756 & 0.744 \\
 & \scriptsize{$\pm$4.1\%} & \scriptsize{$\pm$6.8\%} & \scriptsize{$\pm$3.8\%} & \scriptsize{$\pm$13.2\%} & \scriptsize{$\pm$0.063} & \scriptsize{$\pm$0.050} \\
Stream B only & 77.00\% & 80.10\% & 82.10\% & 72.60\% & 0.763 & 0.752 \\
 & \scriptsize{$\pm$3.7\%} & \scriptsize{$\pm$6.8\%} & \scriptsize{$\pm$3.8\%} & \scriptsize{$\pm$12.3\%} & \scriptsize{$\pm$0.057} & \scriptsize{$\pm$0.036} \\
Always-On & 85.40\% & 89.90\% & 83.50\% & 85.30\% & 0.842 & 0.901 \\
 & \scriptsize{$\pm$0.8\%} & \scriptsize{$\pm$1.1\%} & \scriptsize{$\pm$5.1\%} & \scriptsize{$\pm$3.0\%} & \scriptsize{$\pm$0.016} & \scriptsize{$\pm$0.020} \\
\textbf{SecondOpinion} & \textbf{84.60\%} & \textbf{89.20\%} & \textbf{83.20\%} & \textbf{83.30\%} & \textbf{0.830} & \textbf{0.898} \\
 & \scriptsize{$\pm$0.9\%} & \scriptsize{$\pm$1.7\%} & \scriptsize{$\pm$5.1\%} & \scriptsize{$\pm$5.1\%} & \scriptsize{$\pm$0.026} & \scriptsize{$\pm$0.020} \\
\hline
\end{tabular*}%
}
\end{table}

\subsection{Efficiency and Activation Analysis}

\begin{table}[t]
\footnotesize
\setlength{\tabcolsep}{3.5pt}
\renewcommand{\arraystretch}{0.92}
\centering
\caption{Parameters, FLOPs, and activation rate per configuration.}
\label{tab:efficiency}
\resizebox{\columnwidth}{!}{%
\begin{tabular*}{\columnwidth}{@{\extracolsep{\fill}}lccc@{}}
\hline
\textbf{Configuration} & \textbf{Params (M)} & \textbf{FLOPs (G)} & \textbf{Activation} \\
\hline
Single Stream Only & 4.18 & 0.400 & -- \\
Always-On & 10.68 & 0.855 & 100\% \\
\hline
SecondOpinion -- CXR & 4.78 & 0.442 & 9.23\% \\
SecondOpinion -- VIS & 5.75 & 0.510 & 24.12\% \\
SecondOpinion -- INVIS & 7.15 & 0.608 & 45.71\% \\
\hline
\end{tabular*}%
}
\end{table}

Table~\ref{tab:efficiency} reports per-path Params/FLOPs and activation-weighted averages. SecondOpinion's cost scales with task difficulty, from 4.78M params/0.442 GFLOPs on chest X-ray (9.23\% activation) to 7.15M params/0.608 GFLOPs on invisible fracture (45.71\% activation), 41--71\% of the always-on cost while retaining performance within 0.6--2.2 points of the ceiling. The corresponding AUROC gap between Stream A and the always-on model increased from 1.18 to 8.40 and 15.78 points across the three tasks, indicating that GateKeeper activates Stream B primarily when the additional anatomical information yields the greatest performance gain, particularly under the distribution shift to unseen invisible fractures. Note that reported costs cover the classification pipeline only and exclude the upstream segmentation model, which runs offline prior to inference.

\subsection{Qualitative Analysis}

\begin{figure}[t]
\centering
\includegraphics[width=\columnwidth]{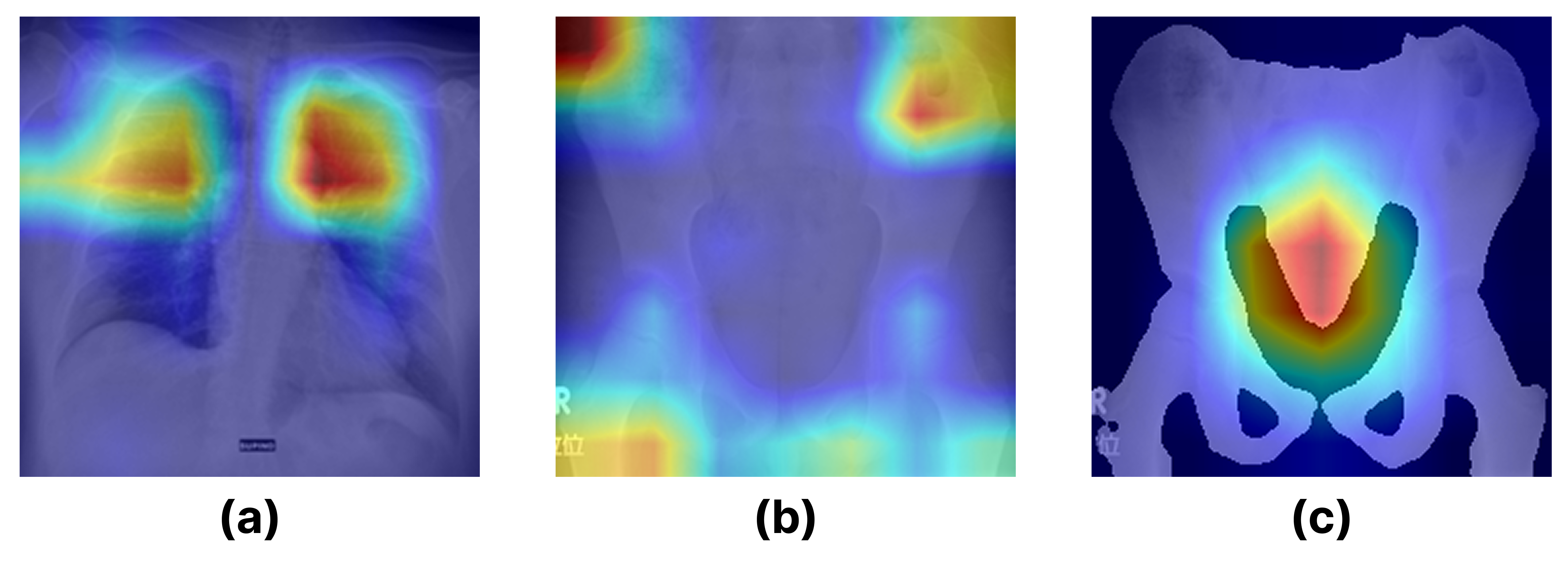}
\caption{GradCAM examples. (a) Chest X-ray (Stream B inactive): Stream A attends correctly to both lungs. (b) Invisible fracture (Stream B active): Stream A attends to irrelevant background corners. (c) Same case post-fusion: attention refocuses onto the bone region.}
\label{fig:gradcam}
\end{figure}

Figure~\ref{fig:gradcam} presents GradCAM visualizations \cite{selvaraju2017grad} for two correctly classified cases. For a COVID-19 chest X-ray case (Stream B inactive), Stream A attends correctly to both lung fields. For an invisible fracture case (Stream B active), Stream A instead attends to irrelevant background corners; after fusion, attention shifts to the central bone region, and the case is correctly classified. This suggests the anatomy-guided second opinion corrects an otherwise poorly localized prediction.

\textbf{Limitations.} The invisible fracture subset contains only 35 cases, so despite reporting 95\% confidence intervals, its point estimates should be interpreted with caution. The pelvic fracture dataset is private, our evaluation uses a single backbone and gating architecture, and our qualitative analysis is limited to two representative cases intended as a sanity check rather than a systematic interpretability evaluation.

\section{Conclusion}

We presented SecondOpinion, a framework that processes every case through a fast primary stream and consults a second, anatomy-guided stream only when GateKeeper judges the primary prediction unreliable. Across two datasets spanning a gradient of diagnostic difficulty, SecondOpinion matches or exceeds prior state-of-the-art performance while activating its anatomy-guided stream on only a fraction of cases. Future work includes extending GateKeeper to additional backbones and modalities, and exploring learned, sample-specific anatomy representations beyond fixed segmentation pipelines.

\section*{Data Use Declaration} The public datasets were obtained from their official repositories and used in accordance with their respective licenses; all required dataset citations are included in the References. The private AMERI PXR dataset was used under IRB approval (No. 2019-1-52) from the Steel Memorial Hirohata Hospital, Japan, and is available from the corresponding author upon reasonable request and with permission from the Steel Memorial Hirohata Hospital, Japan.

\section*{Disclosure of Interests}
The authors have no competing interests to declare.

\bibliographystyle{splncs04}
\bibliography{Paper-0045}
\end{document}